\documentclass[letterpaper, 10 pt, conference]{ieeeconf}  

\IEEEoverridecommandlockouts                              

\usepackage{graphicx}
\usepackage{url}
\usepackage{amsmath}
\usepackage{amssymb}
\usepackage{algorithmic}
\usepackage{algorithm}
\usepackage{amssymb,amsfonts}
\usepackage{color}
\usepackage{float}
\usepackage{listings}
\usepackage[utf8]{inputenc}
\usepackage[english]{babel}
\usepackage[normalem]{ulem}
\usepackage{epstopdf}
\usepackage{booktabs}
\usepackage{cite}
\usepackage{rotating}
\usepackage[colorlinks=true, 
			pdfstartview=FitV, 
			linkcolor=black, 
			citecolor=black, 
			plainpages=false, 
			pdfpagelabels=false, 
			urlcolor=black]{hyperref}
\usepackage[all]{hypcap}
\usepackage{multirow}
\usepackage{xcolor}
\usepackage{colortbl}

\title{\LARGE \bf
How To Not Drive: Learning Driving Constraints from Demonstration}

\author{
        Kasra~Rezaee and Peyman~Yadmellat \\
        \thanks{The authors are with the Noah's Ark Lab., Huawei Technologies, Markham, Ontario, Canada, \protect\url{{kasra.rezaee, peyman.yadmellat}@huawei.com}.} 
}

\begin{document}

\maketitle 
\thispagestyle{empty}
\pagestyle{empty}

\begin{abstract}
We propose a new scheme to learn motion planning constraints from human driving trajectories. Behavioral and motion planning are the key components in an autonomous driving system. The behavioral planning is responsible for high-level decision making required to follow traffic rules and interact with other road participants. The motion planner role is to generate feasible, safe trajectories for a self-driving vehicle to follow. The trajectories are generated through an optimization scheme to optimize a cost function based on metrics related to smoothness, movability, and comfort, and subject to a set of constraints derived from the planned behavior, safety considerations, and feasibility. A common practice is to manually design the cost function and constraints. Recent work has investigated learning the cost function from human driving demonstrations. While effective, the practical application of such approaches is still questionable in autonomous driving. In contrast, this paper focuses on learning driving constraints, which can be used as an add-on module to existing autonomous driving solutions. To learn the constraint, the planning problem is formulated as a constrained Markov Decision Process, whose elements are assumed to be known except the constraints. The constraints are then learned by learning the distribution of expert trajectories and estimating the probability of optimal trajectories belonging to the learned distribution. The proposed scheme is evaluated using NGSIM dataset, yielding less than 1\% collision rate and out of road maneuvers when the learned constraints is used in an optimization-based motion planner. 
\end{abstract}

\section{INTRODUCTION}
Autonomous driving is an active research topic for the last two decades, studied by both academia and industry. An autonomous driving stack is typically divided into multiple modules including localization, perception, and planning. Despite the recent advancements in perception and localization thanks to data-driven approaches, realizing a reliable and safe planning module has remained challenging. The main challenges include i) uncertainty and inaccuracy in predicting other road participants' behavior and upcoming road occupancies; and ii) the diversity of possible scenarios that a self-driving vehicle require to operate in. These challenges often lead to having an overly cautious planning module with a large number of hand-tuned rules and parameters to handle uncertainties and scenario diversities. Additionally, the module needs regular, manual adjustments to deal with novel scenarios. Moreover, there is a reasonable expectation from an autonomous vehicle to exhibit human-like behaviors in order to be predictable for other human road users such as passengers, pedestrians, cyclists, and human drivers, which adds to the complexity of designing an impeccable planning module. 

A planning module typically consists of four components: i) mission planning; ii) behavioral planning; iii) motion planning; and iv) motion control. The missing planning module determines a global route from a start point to a destination to navigate the vehicle. The behavioral planner's role is to make high-level decisions required to follow the mission planner route and traffic rules, and interact with other road users. The motion planner plans a safe and feasible trajectory to implicate the behavioral decisions. 
Lastly, the motion control module control the vehicle to follow the planned trajectories. This paper focuses on behavioral and motion planning as the behaviors of an autonomous vehicle is primely influenced by these two modules. 

A common practice in designing behavioral and motion planners is to processed the detected objects and predicted occupancies in the behavioral planner through a rule-based scheme (\textit{e.g.} state machines and decision trees) and generate a set of costs and constraints for the motion planner to generate a trajectory. For example, if the behavioral decision is to follow a heading vehicle, the decision is converted to constraints on the vehicle speed and the trajectory length to maintain a minimum gap with the heading vehicle. Such a design offers separation of concerns between different modules as the main advantage. However it suffers from several limitations. First, the design is based on an assumption that the predictions and decisions in the upper stream modules are independent from the level level decisions. This is not a valid assumption as in many cases prediction results depends on the vehicle's own behaviors or a behavior decision may need to change if there are no feasible, safe trajectory to follow the planned behavior. Second, the design is not scalable as it requires a preset mapping between all possible behavioral decisions and motion planning constraints. Any new situations require redefining the mapping and analyzing the effect on motion planning when transitioning from one decision to another. Third, the conversion between behavioral decisions and constraints often ignores contextual dependencies and tends to be conservative to ensure safety in as many scenario as possible. Naturally, this may lead to planning failures and/or requiring additional fall-back schemes to relax or prioritize the constraints. For example, a certain distance gap to other objects may be considered as the constraint when passing a parked vehicle, which can work in a majority of scenarios. But it can cause the motion planner to fail in a narrow and tightly-constrained road. A similar situation can be imagined in a merging scenario with a predefined gap and speed.

Recent studies focus on data-driven, learning-based designs to addressed the above-mentioned challenges. Among them are end-to-end designs based on supervised and imitation learning~(IL) methods~\cite{bojarski2016end,schwarting2018planning,yang2018end}, learning a mapping from sensory data to low-level control. End-to-end methods in general require substantial changes in the current autonomous driving stacks, while lacking interpretability and safety assurance. Other approaches include reinforcement learning (RL) and particularly inverse reinforcement learning (IRL) methods~\cite{wu2020efficient,xin2019accelerated}, which involve learning an underlying reward function from human driving trajectories, and using the learned reward function to train a policy. The existing approaches however are simulation-based, which may not generalize well in a real-world setting. More recent approaches~\cite{wei2020perceive,zeng2019end} consider learning a spatio-temporal cost volume for trajectory planning through an end-to-end prediction and planning scheme.

In this paper we look at the behavioral and motion planning from a different perspective, and reformulate the planning problem into two operations of constraint generation and trajectory planning. In most driving tasks, designing a cost function is relatively easy, and the challenge is to define context-dependent constraints for motion planning trajectories. To this end, we propose a new approach to infer motion planning constraints based on human driving data as the main contribution. The inferred constraints are then used in a regular motion planning method to generate trajectories. As the main advantage, the proposed method simplifies the motion planning cost to simple terms to evaluate the dynamic features of a trajectory such as maximum curvature, acceleration, and jerk. The proposed method is also compatible with the existing autonomous driving stacks as it can be considered as an add-on module with minimal or no changes to feed constraints to the existing motion planner modules. Compared to other learning methods, it also enables engineers to adjust motion planning costs without requiring any additional training cycles.

Contributions of this paper can be summarized as
\begin{itemize}
    \item Proposing an approach to employ Variational Auto-Encoder to estimate the probability density of driving demonstrations,
    \item Combining ideas from IRL and maximum margin planning to devise a novel approach to infer constraints from demonstrations in a dynamic environment with continuous states and actions,
\end{itemize}

\section{Related Works}

Raunak~\textit{et al.}~\cite{bhattacharyya2020modeling} consider the human driving problem as a multi-agent, non-linear and stochastic problem with an unknown cost function, and propose multiple modifications to GAIL to solve the formulated driving problem. 
Junning~\textit{et al.}~\cite{huang2019learning} propose a modular framework for adversarial style generative learning with data-driven high-level planning and configurable low-level control. They filter different behaviors into one decision module. The module also supports integration with non-differential
low-level planning or control modules through reparametrized generative adversarial learning.

In~\cite{huang2018learning}, the expert learning is formulated as a max-min problem in the context of RL. The agent chooses a policy and an adversary chooses a reward function that is consistent with the agent’s current knowledge and minimizes the agent’s reward. The goal of the agent is to learn a policy that maximizes the worst possible reward using the ellipsoid method. 

Sacha \textit{et al.}~\cite{rosbach2020planning} use a single reward function for different driving situations. The reward function is computed with path integral MaxEnt-IRL~\cite{Aghasadeghi2011MaximumEI} trained from human driving demonstrations in the context of sampling-based planning algorithm. A similar sampling-based approach is proposed in~\cite{Wu_2020}, where the reward is defined as a linear-structured function with predefined features, whose coefficients are learned from demonstrations. A similar approach~\cite{7535466_swerve} is used for swerving maneuver with static objects on either side of a road. The swerve parameters are learned from human driving demonstrations. 

In \cite{ld_style}, a problem of learning human driving styles on highways is considered. A set of scalar features are extracted similar to a cost function from the expert trajectories. 
Properties such as distance between the vehicles, proper acceleration, and jerk are used to define the features. The cost function is defined as a linear sum of these features. Learning is done in this feature space using MaxEnt IRL, such that the distance between the resulting trajectory and demonstrated trajectory is minimized for a given set of demonstrations. In~\cite{sierragonzalez:hal-01396047}, offline demonstrations are used under MaxEnt IRL for predicting the actions of each of the participating vehicles on a highway scenario. The proposed driver model is formed as a linear feature-based cost function, used to predict the future-time dynamics. Gabriel \textit{et al.}~\cite{kalweit2020deep} propose a hybrid IRL and Q-learning approach, offering faster training compared to other variants of MaxEnt IRL. In~\cite{YOU20191}, both RL and IRL are employed to model multiple driving behaviors. First, an expert agent is trained using RL against a known reward function. Then, the trained agent is used to generate demonstrations. The reward function is extracted from demonstrations using an IRL approach by maximizing the entropy of the joint distribution over a short data segment. 

Above-mentioned studies have mainly focused on learning a reward function (cost function) to imitate expert drivers. While useful, it's unclear how these approaches can be integrated into safety-critical applications such as autonomous driving. In some cases, designing a cost function for motion planning is a relatively easier tasks compared to finding a suitable constraints. As such, few recent work have studied learning constraints from demonstration.

Chou~\textit{et al.}~\cite{Chou_2020} propose a search-based approach to find lower cost trajectories that do not violate the known constraints, as a complementary solution to recover the latent constraints found within expert demonstration shared across multiple tasks. The demonstration set is required to have a boundedly-suboptimal cost that satisfies all the constraints. 
While it is shown that the resulting solution is capable of avoiding collisions in a 2D space, the method is not extensible for more complex constraint representation cases.

An expert agent's behavior is often guided by a set of constraints. In~\cite{scobee2020maximum}, a method is proposed to estimate the underlying constraints using MaxEnt IRL framework. The method is based on finding the frequency of occurring a trajectory in the observed set and iteratively selecting the individual minimal constraint sets with a proper stopping function. 

\section{Constraint Inference}

Consider a deterministic, finite-horizon Markov Decision Process (MDP) $\mathcal{M}$ with constraints represented with tuple $(\mathcal{S},\mathcal{A}, f, g, r, T)$, 
where $\mathcal{S} \in \mathbb{R}^{|\mathcal{S}|}$ is the set of states,
$\mathcal{A} \in \mathbb{R}^{|\mathcal{A}|}$ is the set of actions,
$f: \mathcal{S} \times \mathcal{A} \to \mathcal{S}$ is the state transition function defining next state given current state and action,
$g: \mathcal{S} \times \mathcal{A} \to {0, 1}$ is the constraint on state and action pairs,
$r(s,a): \mathcal{S} \times \mathcal{A} \to \mathbb{R}$ is the scalar reward for taking action $a$ in state $s$,
and $T$ is the decision time horizon. 
An agent behaving optimally in this environment is effectively solving the optimization problem defined below.
\begin{equation}
	\begin{aligned}
		\max_{a_t} \quad & \sum_{t=0}^{T}{r(s_t, a_t)} \\
		\textrm{s.t.} \quad & s_{t+1} = f(s_t, a_t), & t=0 \dots T-1 \\
		                    & g(s_t, a_t) = 0,       & t=0 \dots T-1 \\
	\end{aligned}
	\label{eq_optim}
\end{equation}

In a regular RL problem, the MDP $\mathcal{M}$ is fully known and the goal is to solve the optimization problem in \eqref{eq_optim} to find the sequence of actions $a_0, \dots , a_{T-1}$. 
In this work, we assume all the elements of $\mathcal{M}$ are known except the constraint function $g$,
and the goal is to infer $g$ by observing the demonstrations from an expert that has optimized \eqref{eq_optim}.
Suppose we have a set of $n$ demonstration instances ${\mathcal{D} =\{(\mathsf{S}^i,\mathsf{A}^i,r^i)\}_{i=1}^n}$ sampled from an expert behaving optimally in MDP that is maximizing $\mathcal{M}$,
where $\mathsf{S}^i$ and $\mathsf{A}^i$ are lists of length $T$ of the states and actions, respectively, observed by the expert.
The reward function $r^i$ is assumed to be unique for each instance and correspond to the experts goal for that instance.

We develop our approach on top of the foundation described in \cite{scobee2020maximum} to find the constraint function from demonstrations.
At a high level, the approach presented in \cite{scobee2020maximum} can be summarized into starting with an empty constraint set and iteratively performing the following steps:
\begin{itemize}
	\item For each demonstration $i$, find $\hat{\mathcal{D}}^i$, the solution to the optimization problem \eqref{eq_optim} considering the latest estimate of the constraint function $g$ with the initial state $s_0$ being the first state from that demonstration instance.
	\item Count the state-action visitation of $\hat{\mathcal{D}}^i$ for all demonstrations, and find the state-action pair with the highest visitation that is not present in expert demonstrations.
	\item Add state-action pair from previous step to the set of state-actions constraints.
\end{itemize}
This approach relies on counting and comparing the visitation, thus, it is only applicable discrete states and action with a stationary environment with fixed start and goal.
Given the nature of driving problems, we devise an algorithm that is applicable to dynamic environments with continuous states and actions and varying start and goal states.
We train a model to match the probability distribution of the expert demonstration instances.
Effectively, this model would estimate the probability that a given state-action pair is similar to the ones in the expert demonstrations.
Additionally, instead of finding the state-action with highest visitation, we employ a sample-based approach to find suitable constraints. 
We utilize an approach similar to Maximum Margin Planning \cite{ratliff2009learning} to update the constraint function $g$. 
The steps of our proposed solution are detailed in Algorithm~\ref{alg:cap}.

\begin{figure}
	\centering
	\includegraphics[width=1.0\columnwidth]{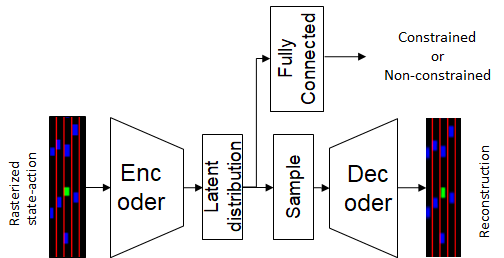}
	\caption{The architecture of the constraint function}
	\label{fig_architecture}
\end{figure}

\begin{algorithm}
	\caption{Constraints inference algorithm.}\label{alg:cap}
	\begin{algorithmic}
		\STATE $p(.) \gets $ probability distribution function of $\mathcal{D}$
		\STATE Initialize the constraint model $g(.)$
		\WHILE{ not converged }
			\STATE $\hat{\mathcal{D}}^i \gets$ solution of \eqref{eq_optim} for $\mathcal{D}^i$
			\STATE $s_t, a_t \in \mathcal{D}^i \gets $ no-constraint label
			\STATE $\hat{s}_t, \hat{a}_t \in \hat{\mathcal{D}}^i$, where $p(\hat{s}_t, \hat{a}_t) < p_{th} \gets $ constraint label
			\STATE Update $g(.)$ with gradient descent with labeled inputs $s_t, a_t$ and $\hat{s}_t, \hat{a}_t$
		\ENDWHILE
	\end{algorithmic}
\end{algorithm}

\subsection{Constraint Inference in Autonomous Driving}
Commonly, the IRL approaches use model-free RL for the optimization step (\textit{e.g.} see \cite{fu2017learning,malik2021inverse}), which are computationally expensive. 
However, in the context of motion planning for autonomous driving, there are efficient and effective solutions available for the optimization step.
Therefore, instead of relying on RL approaches for the optimization step, we employ a sample-based motion planner to find the sequence of actions that optimize \eqref{eq_optim}.

We use a sample-based approach similar to the one used in \cite{casas2021mp3} to solve the motion planning problem. 
For a given starting state corresponding to a demonstration sample, we generate a number of trajectories in the Frenet frame (road coordinate frame) with varying speed and lateral positions. 
Then, the trajectories are checked against the constraint function $g$, and the ones that violate $g$ are discarded. 
Finally, the reward corresponding to the remaining trajectories are calculated using function $r_i$. A trajectory with the highest reward is the solution to the optimization problem.

\subsection{Implementation}
We utilize a Variation Auto-Encoder (VAE) for the probability density function $p(.)$.
When trained properly, VAE will have lower reconstruction error for samples from training dataset compared to samples that are outside the training distribution.
Therefore, the reconstruction error can be used to identify state-action pairs that are different from expert demonstrations. 

The constraint function $g$ is represented with a neural network, trained as a classifier to distinguish between constrained and non-constrained state-action pairs. 
The input to the neural network is state and action and the output is whether they belong to the constrained or non-constrained class. 
We employ a VAE as the architecture for the function $g$ with an additional fully connected section from the latent variables for the classifier as shown in Figure~\ref{fig_architecture}. 
While the main training cost for the classifier is based of the constrained and non-constrained labels, the model also has an auxiliary cost associated with VAE latent distribution and reconstruction. 
These auxiliary costs help with generalizing the training. 
For a given demonstration instance $\mathcal{D}^i$, the loss for training would be
\begin{equation}
	\begin{aligned}
		\mathcal{L}^i = & -\sum_{t=0}^{T}{\ln{(1 - y_t^i)} + \ln{\hat{y}_t^i}} \\
		                & + \sum_{t=0}^{T}{\text{RMSE}(x_t^i,\tilde{x}_t^i)}   \\
		                & + \sum_{t=0}^{T}{KL(z_t^i,\mathcal{N} (0,I))},       
	\end{aligned}
	\label{eq_cost}
\end{equation}
where $y_t^i$ and $\hat{y}_t^i$ are the classifier outputs for station-action pairs from demonstration instance and optimization solution, respectively,
$x_t^i$ and $\tilde{x}_t^i$ are the VAE input and its reconstruction representing state-action pairs from demonstration instance, and $z_t^i$ is the latent variable distribution for station-action pairs from demonstration instance. 
The auxiliary cost of the VAE are only limited to the state-action pairs from the demonstrations and state-action pairs from the optimization solution is not included, as they might not be valid inputs.

We encode state and action with 2D multi-channel images in the form of dynamic Occupancy Grid Maps (OGM).
A dynamic OGM includes one channel for the occupancy and two channels for speed of occupied pixels along the $x$ and $y$ coordinates.
All dynamic OGMs are ego-centric. 
We use one dynamic OGM for the positions of the ego to represent the ego action.
Another dynamic OGM will include position and speed of other vehicles around ego representing the state. 
Additionally, another single channel will represent the road markings in the form of a 2D image.
Since the OGM's are ego-centric, the ego is always at the center, and its action will result in other vehicles moving in the 2D image.

\subsection{NGSIM Dataset and State Transition Function}
NGSIM dataset is an open dataset tracking all the vehicles in a road section using overhead cameras.
Vehicles are tracked as long as they are in the visible range of the cameras.
We generate a demonstration instance for each vehicle and frame combination in the dataset that the vehicle would be available for at least T seconds after that frame.
For a given vehicle and frame, the vehicle would be the ego, and its sequence of positions for the next $T$ seconds represent its actions for that demonstration instance.
Positions of other vehicles around the ego in a frame would make the state of the environment for ego in that frame.

We need a state transition function for the optimization step, to calculate the state of the environment for new actions not in the dataset.
Ideally, this would be a simulator that can predict the reaction of other vehicles to movements of ego.
In the absence of access to such simulator, we employ an approach similar to the one in \cite{bhattacharyya2018multi}.
We assume that other vehicle will continue with their movement irrespective of the ego movements.
After the model is trained, the expectation is that the solution of the optimization problem would be close to the expert demonstrations, therefore the assumption that other vehicles will continue with their movement would be valid.

\section{Experiments and Result}
We have trained and evaluated the proposed approach on the US101 potion of the NGSIM dataset.
The vehicles positions in local coordinate (the road coordinate system) was used to generate dynamic OGMs.
OGMs were generated with a resolution of 0.5 meter per pixels with width and height of 32 by 128, resulting in 16 meters of lateral coverage and 64 meters of longitudinal coverage with the ego being at the center of the OGM.
The state variable consists of one dynamic OGM and an extra channel for lane markings, resulting in a 4x32x128 variable.
The action, corresponding to the ego position and speed is also a dynamic OGM with the size 3x32x128.
The two variables are concatenated and yield the input to various models.

\subsection{Probability Distribution of the Demonstrations}
For VAE architecture, we employed a Convolution Neural Network for the encoder side, normal distribution with mean and variance for the latent variable, and a Convolution-transpose Neural Network for decoder.
A cyclical schedule for $\beta$ \cite{fu2019cyclical} was employed to train the VAE.

\begin{figure*}
	\centering
	\includegraphics[width=1.0\columnwidth]{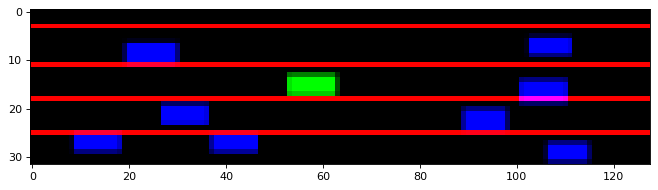}
	\includegraphics[width=1.0\columnwidth]{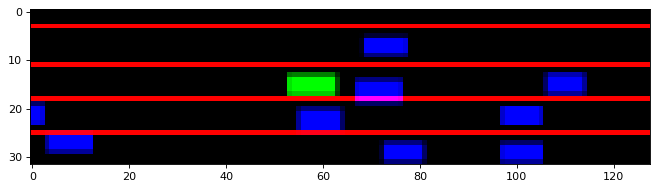}
	\includegraphics[width=1.0\columnwidth]{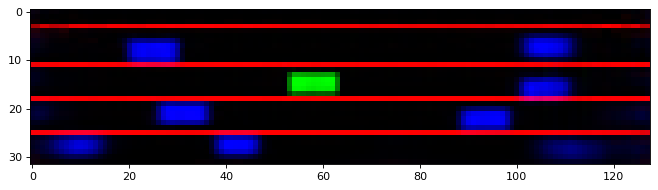}
	\includegraphics[width=1.0\columnwidth]{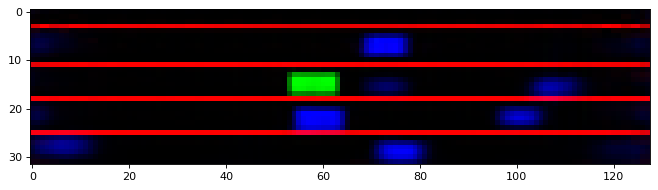}
	\caption{VAE input (left) and reconstruction (right) for a sample demonstration dataset (top) and a sample with ego too close to other vehicles (bottom)}
	\label{VAE}
\end{figure*}

Figure~\ref{VAE} shows the input and reconstruction for two input samples.
The left image is a sample directly from the demonstration data and its reconstruction shows vehicles in their respective locations.
The right image is an artificially created sample with the ego too close to another vehicle.
We can see that the reconstruction does not include the vehicle that is too close to ego, which will result in relatively higher reconstruction error.

\subsection{Motion Plan for Optimization}
To find the optimal motion plan for a demonstration instance given the constraint function, we used a sample based motion planner in the road frame.
A total of 91 candidate trajectories, combination of 7 lateral positions and 13 longitudinal speeds, were generated.
The 7 lateral positions cover the range from center of the lane to the left of ego to the center of the lane to the right of ego.
The 13 longitudinal speeds were evenly spaced from 0 to 24 [m/s].
The trajectories that satisfied the constraint function were sorted according to a cost function that minimized lateral and longitudinal jerk, with speed as close to speed limit as possible, and lateral position close to the ego lane 5 seconds into the future.

\subsection{Constraint Function}
\begin{figure}
	\centering
	\includegraphics[width=1.0\columnwidth]{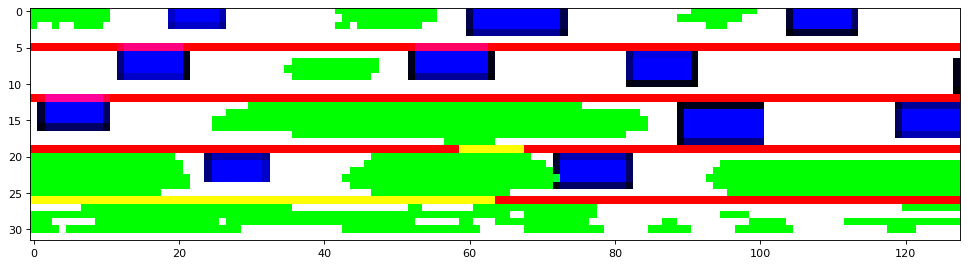}
	\caption{The drivable region identified by the constraint function. 
		The green pixels show the positions for ego that are not constrained. 
	This figure is generated by evaluating the constraint function for different ego actions (ego positions) in this current state}
	\label{constraint_func}
\end{figure}

The constraint function has an architecture similar to the VAE with an additional fully-connected section for the classifier part. 
We used the pretrained VAE weights as the starting point for the training of the constraint model.
Figure~\ref{constraint_func} show the result of the constraint function in the form of a drivable area. 

Once the training of the constraint function is concluded, the constraint function can be used in the optimization problem of a motion planner.
We evaluated the resulting motion planner with samples from the NGSIM dataset and checked the performance of the motion planner system.
Given that the simplified transition model employed does not simulate the reaction of other vehicles to the movements of the ego vehicle, we limited the evaluation to a single horizon and did not attempt to perform a rolling horizon simulation.
A summary of the performance of the resulting motion planner is presented in Table~\ref{mop_table}.
Overall, the constraints were effective in preventing collisions and keeping the vehicle withing the road bounds.
However, this result was achieved by having rather aggressive constraints, which resulted in the motion planner to fail to find a valid trajectory that satisfy the constraints in 1.5\% of cases.

\begin{table}[h!]
	\begin{center}
		\caption{Summary of the performance of the motion planning solution utilizing the constraint model learned from demonstrations}
		\label{mop_table}
		\begin{tabular}{c|c|c} 
			\textbf{Collision} & \textbf{Out of road} & \textbf{No MoP solution} \\
			\textbf{\%}        & \textbf{\%}          & \textbf{\%}              \\
			\hline
			                   &                      &                          \\
			0.5                & 0.3                  & 1.6                      \\
		\end{tabular}
	\end{center}
\end{table}

\section{CONCLUSIONS \& FUTURE WORK}

Motion planning algorithms that optimize simple cost function have the benefit of being interpretable and easily tunable.
However, the resulting trajectories can be unnatural in edge and extreme cases. 
In the literature, the common approach to overcome this limitation is using more complex and general cost functions tuned using real driving demonstrations.
But this result comes at the expense of interpretablity and ease of tuning the cost function.
In this work we presented an approach to define the driving constraints from demonstrations instead of the cost.
The cost function would remain simple to maintain its benefits, while the constraints limit the extent of the movement so that the resulting trajectory remains close the movements observed from the real demonstrations. 
We trained a constraint function with a VAE architecture at its core using the NGSIM dataset.
The constraint function was able to steer the ego vehicle away from collisions and out of road regions.
However, the resulting constraints were rather restrictive as there were considerable cases where the motion planner optimization failed to find a solution that satisfy the constraints.

In this work, we encoded the state and actions with dynamic OGMs. 
In our future work, we plan to directly use vector features from the environment and leverage graph neural network (GNN) to map from inputs to constraints.

\addtolength{\textheight}{-7cm}   

\bibliographystyle{IEEEtran}
\bibliography{IEEEabrv,bibliography} 
                                  
\end{document}